# Breaking BERT: Gradient Attack on Twitter Sentiment Analysis for Targeted Misclassification

Akil Raj Subedi[1] Taniya Shah[1] Aswani Kumar Cherukuri[1*] Thanos Vasilakos[2]

[1]School of Computer Science Engineering and Information Systems, Vellore Institute of Technology, Vellore, India

[2]Center for AI Research (CAIR), University of Agder(UiA) , Grimstad, Norway

[2]College of Computer Science and Information Technology, IAU, Saudi Arabia

(*Corresponding Author: cherukuri@acm.org)

## Abstract

Social media platforms like Twitter have increasingly relied on Natural Language Processing (NLP) techniques to analyze and understand the sentiments expressed in the user-generated content. One such state-of-the-art NLP model is Bidirectional Encoder Representations from Transformers (BERT) (Devlin, Chang, Lee and Toutanova, K. 2018) which has been widely adapted in sentiment analysis. BERT, however, is susceptible to adversarial attacks. This paper aims to scrutinize the inherent vulnerabilities of such models in Twitter sentiment analysis. It aims to formulate a framework for the construction of targeted adversarial texts capable of deceiving these models; and that while maintaining stealth. In contrast to conventional methodologies, such as Importance Reweighting, this framework's crux resides in its reliance on gradients to prioritize the importance of individual words within the text, i.e., it uses a white-box approach to attain fine-grained sensitivity, pinpointing words that exert maximal influence on the classification outcome. This paper is organized into three interdependent phases. It starts with fine-tuning a pre-trained BERT model on Twitter data. It then analyzes gradients of the model to rank words on their importance, and iteratively replaces those with feasible candidates until an acceptable solution is found. Finally, it evaluates the effectiveness of the adversarial text against the custom-trained sentiment classification model. This assessment would help in gauging the capacity of the adversarial text to successfully subvert classification without raising any alarm.

**Keywords**: Gradient Attack, Sentiment Analysis, BERT, Twitter, Adversarial Text

## 1 Introduction

Adversarial machine learning has emerged as a challenging area of study, focusing on the manipulation of input data to expose vulnerabilities in machine learning models. Within this context, we undertake a comprehensive exploration of adversarial attacks on the Bidirectional Encoder Representations from Transformers (BERT), a specific application to Twitter sentiment analysis. Our research centers on sentiment analysis, which involves categorizing text data based on sentiment. This approach offers valuable insights into user emotions, opinions, and attitudes, particularly in the context of platforms like Twitter.

Machine learning models typically operate under the assumption that training and testing data originate from a consistent and benign statistical distribution. In practice, however, real-world scenarios introduce adversarial data that challenges these assumptions, resulting in unreliable model outputs and exposure to vulnerabilities. At the heart of our study lies the concept of "adversarial examples." These are thoughtfully constructed to subtly alter clean text samples that would typically be correctly classified by machine learning models, thereby deceiving the model into making incorrect predictions while remaining imperceptible to human observers.

BERT being good at understanding the context of short and informal texts makes it a suitable choice for tweet sentiment analysis. BERT can also learn from what it already knows through transfer learning which means that it does not have to be trained again when faced with new domains or languages, hence saving time. This flexibility, coupled with its state-of-the-art performance in natural language processing tasks, positions BERT as a powerful tool in sentiment analysis. By utilizing various dilation rates and integrating a sentic knowledge base, BERT can effectively capture long-term dependencies and perform concept-level sentiment analysis, further solidifying its position as a leading choice in sentiment analysis tasks, particularly in the context of social media data analysis like tweet sentiment analysis. However, it's essential to acknowledge that BERT is not immune to vulnerabilities. As highlighted in a research (Li, Ma, Guo, Xue, and Qiu 2020), BERT can be susceptible to perturbations that consider the surrounding context.

The success of such attacks depends on three crucial utility-preserving properties: maintaining human prediction consistency, ensuring semantic similarity, and upholding language fluency. These properties dictate that, despite manipulation, human predictions should remain unaltered, the meaning of the content should endure as perceived by humans, and the generated examples should appear natural and grammatical. Nevertheless, prior methods in this domain often struggle to fulfill all three requirements concurrently. Techniques like word misspelling (Gao, Lanchantin, Soffa and Qi 2018), single-word erasure (Li, Monroe, and Jurafsky 2018), and phrase insertion/removal (Papernot, McDaniel, Goodfellow, Jha, Celik and Swami 2017) have been used to yield sentences that deviate from natural fluency and coherence as pointed out in another paper(Jin, Jin, Zhou and Szolovits 2020).

In our research, we introduce an innovative approach, building upon the TEXTFOOLER framework (Jin, Jin, Zhou and Szolovits 2020), tailored for white-box sentiment analysis with BERT in the context of Twitter sentiment analysis. Unlike the black box setting, where model architecture and parameters are concealed, our method capitalizes on access to model internals to calculate word importance using gradients. Our approach centers on the creation of semantically and syntactically similar adversarial examples that adhere to the three fundamental criteria: consistency in human predictions, semantic alignment, and linguistic fluency. We keep our interest limited to misclassifying the negative sentiment texts. Specifically, we identify pivotal words for the target BERT model and prioritize their replacement with semantically analogous and grammatically accurate alternatives until the model's prediction is either modified or undermined to lower confidence. We further enforce the robustness of our approach by addressing the task of handling slang and misspelled words before figuring out candidate words; this step is necessary for our specific requirement of tweets analysis.

Our motivation for this work lies in critically examining the robustness of sentiment analysis models as robust as BERT. While these models are widely used for analyzing public sentiment, their susceptibility to adversarial attacks and manipulation raises concerns about their reliability and robustness. Our goal is to challenge the prevailing assumption of model robustness and highlight the need for more sophisticated and resilient sentiment analysis techniques.

Our report is divided into several sections. First, we introduce some technical background on the topic and back it up with a literature survey. Then we present the security issues in our theme, where we dive deeper into the potential issues and vulnerabilities involved in a white-box scenario of BERT; we use our framework to perform adversarial attack on the model and investigate the challenges associated with this approach. Then we discuss the broader security implications and consequences of our chosen theme and consider how the security issues identified can impact the wider context of natural language processing and machine learning security. We then analyze in detail the previous two sections and conclude our project.

## 2 Background

The success of adversarial attacks on sentiment analysis models like BERT hinges on three key factors. Firstly, maintaining human prediction consistency is crucial. This means making changes to the text that lead the model to a different prediction while not altering how humans interpret the sentiment significantly. Balancing this requires careful adjustments to keep the manipulated text inconspicuous to human readers. Secondly, ensuring semantic similarity is vital. This involves modifying the text in a way that preserves the original meaning and intent. Lastly, upholding language fluency is essential. This ensures that the manipulated text appears grammatically correct and coherent. This linguistic coherence is crucial to avoid drawing attention to the manipulated text. The effective interplay of these three factors is at the core of successful adversarial strategies.

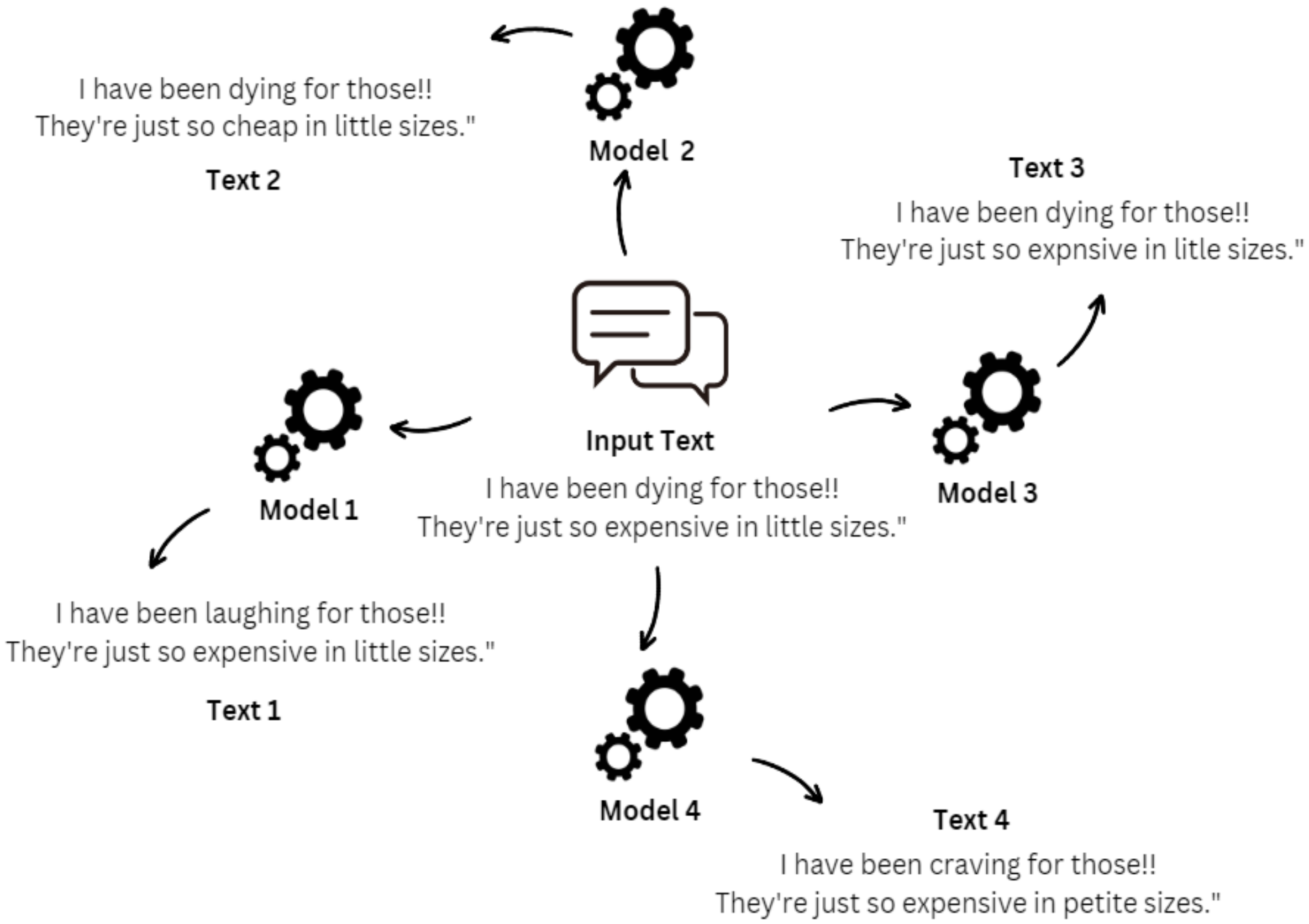

**Fig. 1** Adversarial Text Examples

Fig. 1 illustrates the generation of adversarial text examples for the input text "I have been dying for those!! They're just so expensive in little sizes." Four variations are presented, each with a distinct linguistic alteration. Text 2 fails to maintain human prediction consistency as the replacement of "expensive" with "cheap" induces a significant shift in sentiment. Text 1 deviates from the source's semantic structure, replacing "dying" with "laughing," disrupting the original meaning. Text 3 compromises language fluency by introducing a spelling error. In contrast, Text 4 successfully fulfills all requirements by substituting "dying" with "craving," "little" with "petite," while preserving positive sentiment and ensuring linguistic fluency. This demonstrates the challenges in crafting adversarial examples that adhere to human prediction consistency, semantic similarity, and language fluency simultaneously.

Several attempts have been made to misguide BERT. In the study (Gao, Lanchantin, Soffa and Qi 2018), the authors propose DeepWordBug, an algorithm designed for inducing misclassification through subtle text perturbations in a black-box setting. Applying this method to Enron spam emails and IMDB movie reviews, the results demonstrate a substantial reduction in classification accuracy from 99% to 40% for Enron and from 87% to 26% for IMDB, underscoring the effectiveness of DeepWordBug in generating impactful adversarial examples. However, the approach compromises spelling accuracy due to character-level changes, posing challenges to language fluency in the context of adversarial attacks.

Another work (Li, Monroe and Jurafsky 2016) presents a methodological framework for analyzing neural model decisions by systematically erasing different components of the model representation. The methodology spans erasure of input word-vector dimensions, intermediate hidden units, and input words, providing insights into decision-making processes. While the paper offers several approaches for evaluating the impact of erasure, from metrics assessment to reinforcement learning-based deletion of minimum input words to flip model decisions, it falls short of fully adhering to the crucial utility-preserving properties. Notably, the erasure of input words introduces challenges to language fluency, as passages describing different aspects or overall sentiment may be deleted, impacting the model's interpretability negatively.

A paper (Papernot, McDaniel, Goodfellow, Jha, Celik and Swami 2017) introduces a black-box attack strategy, enabling an adversary to exert control over a remotely hosted DNN. The attack involves training a local substitute model, generating synthetic inputs labeled by the target DNN, and using this substitute to craft adversarial examples. The study demonstrates the effectiveness of this approach by attacking DNNs hosted by MetaMind, Amazon, and Google. The misclassification rates are notably high, with MetaMind misclassifying 84.24%, Amazon at 96.19%, and Google at 88.94%. Despite achieving successful misclassification, it is crucial to note that the synthetic nature of the adversarial examples raises concerns about language fluency, which may affect the interpretability of the model's decisions in practical scenarios.

Likewise, several successful attempts (Liu et al. 2022; (Hofer, Schöttle, Rietzler and Stabinger 2021; Mary, Sudhan and Narayanan 2022) have been made to attack BERT by performing character level substitutions. While they manage to degrade the performance of the model, they are doing so at the expense of overlooking aforementioned criteria.

In contrast to prior works, TEXTFOOLER (Jin, Jin, Zhou and Szolovits 2020) stands as a significant advancement, distinguishing itself by addressing the crucial requirements mentioned above. The methodology not only identifies pivotal words for the target model but also prioritizes replacements with semantically analogous and grammatically accurate alternatives, showcasing a refined and nuanced approach. It achieves this by making substitutions only when the replaced word is similar to the target word, belongs to the same part of speech, and the generated adversarial example maintains semantic similarity with the original text. Adversarial examples crafted by TEXTFOOLER demonstrate the capability to reduce the accuracy of almost all target models in various tasks to below 10%, utilizing less than 20% of the original words, i.e., it can generate adversarial examples with minimal perturbations. This work firmly establishes TEXTFOOLER as a potent tool, maintaining a harmonious balance between model misdirection and linguistic fidelity. In our work, we follow in its footsteps.

## 3 Literature Survey

### 3.1 Sentiment Analysis

Sentiment analysis focuses on classifying text into primary sentiments. Organizations can leverage sentiment analysis to comprehend customer sentiments and attitudes towards their products and services, leading to improvements in customer satisfaction, brand reputation, and revenue. A paper (Tan, Lee and Lim 2023) discusses the diverse applications of sentiment analysis, such as its use in political analysis to gauge public opinion on political matters and in the financial industry for predicting stock prices and identifying investment opportunities. Furthermore, it provides an overview of recent advancements in sentiment analysis techniques, including preprocessing, feature extraction, and classification methods, along with insights into challenges, limitations, and future research prospects in this field.

Twitter Sentiment Analysis (TSA) has received special attention due to the platform's widespread use for expressing opinions. Despite its importance, sentiment analysis wasn't always an easy task to perform. A research (Zimbra, Abbasi, Zeng and Chen 2018) suggests that the then TSA approaches struggled to achieve high accuracy, often falling below 70%. The research explores two key motivations for TSA: gaining insights into business and social issues and improving TSA techniques. TSA faces challenges like tweet brevity, sentiment class imbalance, and stream-based tweet generation, addressed through various techniques such as sentiment information propagation and stream-based classifiers. A benchmark evaluation of 28 systems across Twitter datasets reveals overall lackluster performance, with an average accuracy of 61%. Domain-specific approaches outperform general-purpose ones, but challenges persist in detecting sentiment expressions. An error analysis identifies 13 error categories, shedding light on common classification mistakes. Finally, the study applies select systems to event detection, showing that systems with better sentiment class recall enhance event detection capabilities.

### 3.2 Machine Learning in Sentiment Analysis

A study (Alshammari and AlMansour 2019) investigates the distinctive challenges associated with Twitter sentiment analysis and performs a comprehensive review of the literature to assess how prevailing approaches tackle these challenges. The research incorporates a benchmark assessment encompassing 28 notable academic and commercial systems, evaluating the classification of sentiment in tweets across five diverse datasets. Utilizing an error analysis, the article pinpoints prevalent causes of classification inaccuracies. Additionally, chosen systems are employed in an event detection case study, broadening the evaluation's scope. In conclusion, the paper summarizes essential trends and insights, providing valuable recommendations to shape the future trajectory of approaches in Twitter sentiment analysis.

Some authors (Abbas, Memon, Jamali, Memon and Ahmed 2019) focus on performing sentiment analysis using a combination of statistics, natural language processing, and machine learning techniques to extract and categorize the emotional content from textual data. The study employs the multinomial Naive Bayes (MNB) classification algorithm to classify movie reviews based on their overall sentiment, distinguishing between positive and negative sentiments. The methodology involves representing each document as a "bag" of words, wherein the word order is disregarded, and emphasis is placed on the frequency of each word. This "bag" of words serves as the basis for training the MNB classifier on the dataset, and the model's performance is evaluated using a holdout set. Remarkably, this approach significantly enhances text categorization performance, achieving an impressive accuracy of approximately 90%, even when dealing with a substantial dataset.

A paper (Singh and Tripathi 2021) sources a dataset from Kaggle, specifically centered around challenges in Artificial Intelligence presented by KFC and McDonald's. It is an extensive collection of over 14,000 tweets. The researchers clean the dataset using the Term Frequency-Inverse Document Frequency (TF-IDF) technique. Once the data is prepared, the authors conducted testing with three distinct classification algorithms: Support Vector Machine (SVM), Random Forest (RF), and Decision Tree (DT). Notably, the Decision Tree algorithm emerged as the top performer, with an accuracy rate of 88.51%.

Focusing on tweets that include text, emojis, and emoticons, a group of researchers (Arumugam, Snegaa and Jayanthi 2023) employ four algorithms—Multinomial Naive Bayes (MNB), Random Forest, Support Vector Machine (SVM), and Decision Tree—for sentiment categorization. To enhance accuracy, predictions from these algorithms are consolidated using the Voting Classifier, an ensemble learning technique. Data preprocessing involves techniques like stop word removal, stemming, and conversion of emojis and emoticons into text representations. Each algorithm's performance is evaluated on the preprocessed data using metrics such as accuracy, precision, F1-score, and recall. SVM outperformed other algorithms with 96.27% accuracy. Additionally, ensemble learning techniques like bagging are employed to boost overall performance, and the Voting Classifier further improved accuracy to 97.21%.

We have reviewed only classical models so far. However, there are more robust and fine-grained approaches to sentiment analysis that have been introduced gradually in the background. There are also models that are specifically tailored for sentiment analysis. A paper (Tymann, Lutz, Palsbröker and Gips 2019) introduces VADER (Valence Aware Dictionary and sEntiment Reasoning), a rule-based model for sentiment analysis tailored to the challenges posed by social media text, particularly microblogs like Twitter and Facebook. VADER combines five generalizable rules that represent syntactic and grammatical conventions for expressing sentiment intensity with a gold-standard lexicon of lexical features specifically tuned to microblog sentiment. Notably, the study discovers that VADER matches ground truth sentiment intensity more accurately than individual human raters, with an astounding correlation coefficient of $r = 0.881$. Moreover, VADER outperforms human raters who were able to classify tweets into positive, neutral, or negative categories with an F1 score of 0.96, showing superior classification accuracy. The simplicity, extensibility, and generalizability of VADER across various domain contexts are among its benefits.

On a similar note, another study (Elbagir and Yang 2019) explores sentiment analysis of Twitter data, with a focus on multi-classification using the Valence Aware Dictionary for sEntiment Reasoner (VADER). In contrast to earlier studies that focused primarily on binary classification, this work explores sentiment analysis in ternary and multiple classes, specifically within the framework of tweets pertaining to the US election of 2016. The results show a noteworthy sentiment distribution: roughly 29% of tweets expressed positivity, 22.89% expressed negativity, 46.7% expressed neutrality, and a smaller percentage expressed very negative sentiments. The prevailing neutral sentiments are due to an imbalance in data and the application of a broad vocabulary for the classification of political content.

VADER is clearly among the top options for sentiment analysis however it cannot be fine-tuned to match datasets or domains. Unlike VADER, there are robust models like BERT, which has flexibility and contextual awareness, so continues to have an advantage in sentiment analysis. A paper (Jain, Quamer, Saravanan, and Pamula 2023) introduces a novel BERT based Dilated Convolutional Neural Network (BERT-DCNN) model, utilizing BERT as a pre-trained language model for generating word embeddings. It yields an accuracy of 87.1% when applied to Twitter airline sentiment data. The model incorporates three parallel layers of Dilated Convolutional Neural Network (DCNN), complemented by a global average pooling layer, facilitating effective fine-tuning. Notably, the BERT-DCNN model excels in dimensionality reduction while accommodating an increase in related dimensions without sacrificing information. The utilization of various dilation rates enables the model to capture long-term dependencies effectively. Additionally, the integration of a sentic knowledge base empowers the model for concept-level sentiment analysis.

There are other strong cases for BERT. Another paper (Pota, Ventura, Catelli, and Esposito 2020) illustrates a case study applying this approach to Italian, including a comparative analysis with existing Italian solutions. This work proposes an innovative approach to Twitter sentiment analysis by addressing the challenges posed by the platform's distinctive jargon. The method involves two key steps: firstly, transforming tweet jargon, including emojis and emoticons, into plain text using language-independent or easily adaptable procedures. Secondly, the classification of resulting tweets is performed using the BERT language model, pre-trained on plain text instead of tweets. This approach offers two advantages: it leverages readily available pre-trained models in multiple languages and taps into larger plain text corpora, surpassing the limitations of tweet-only datasets. The obtained results underscore the efficacy of the proposed method and suggest its promise for other languages, emphasizing its methodological versatility.

### 3.3 Adversarial Attack on BERT

Adversarial attacks on BERT are difficult as (Hauser, Meng, Pascual and Wattenhofer 2021) shows. The study investigates attacks that rely on four-word substitutions and performs a thorough analysis that uses probabilistic and human evaluation. Remarkably, the results show that 96% to 99% of these attacks are unable to maintain semantics, suggesting that the main factor determining their success is providing the BERT model with poor quality data. In response, the paper presents a novel defence mechanism that includes post-processing and data augmentation steps and drastically lowers the success rates of state-of-the-art attacks to less than 5%. This examination contradicts the widely held belief regarding BERT's susceptibility to adversarial attacks, highlighting its enhanced resilience to such attacks.

This is in accordance with what we have discussed in the background section – most attacks do not adhere to the constraint that despite manipulation, human predictions should remain unaltered, the meaning of the content should endure as perceived by humans, and the generated examples should appear natural and grammatical. Some works do overcome this.

A paper (Garg, and Ramakrishnan 2020) uses black-box attack for generating adversarial examples in text classification. The authors have proposed BAE (BERT-based Adversarial Examples) which accomplishes token replacement and insertion in the source text using contextual perturbations from a BERT masked language model (MLM). Evaluating different datasets and models like wordLSTM and wordCNN, BAE attacks are more effective which achieves significant drops of 40-80% than other baseline attacks in test accuracies.

Another paper (Li, Ma, Guo, Xue and Qiu 2020) presents BERT-Attack, an efficient method for creating adversarial samples using pre-trained masked language models like BERT. The strategy involves turning BERT against its fine-tuned models and other deep neural models in downstream tasks, effectively leading target models to make incorrect predictions. BERT-Attack outperforms existing attack strategies, achieving higher success rates and lower perturbation percentages. Crucially, the generated adversarial samples maintain fluency and preserve semantics. The after-attack accuracy averages below 10%, indicating successful perturbation of most samples to mislead state-of-the-art classification models. Furthermore, the perturb percentage is less than 10%, representing a substantial improvement over previous approaches.

Also, there is a technique known as the Modified Word Saliency-based Adversarial Attack (MWSAA)

(Waghela, Rakshit and Sen, J. 2024). The MWSAA misleads classification models by strategically modifying input texts to confuse them while maintaining semantic coherence through refining classical approaches and incorporating word saliency and semantic similarity metrics. Comparing with existing adversarial attack techniques, it is evident that both attack success rate and preservation of text coherence are far much better in MWSAA than any other method. Furthermore, the proposed changes such as embedding contextual information and imposing semantic consistency constraints all contribute to its improved effectiveness in evading detection by classifiers making it a potential means for shoring up natural language processing defenses against adversarial attacks. This means that a smaller number of words are replaced when MWSAA is used compared to PWWS implying that fewer substitutions are needed for successful attacks to be launched with high efficacy. Different victim language models' data sets included in this program produced an average attack success rate ranging from 42.50% to 95.50%.

There is another case (Coalson, Ritter, Bobba and Hong 2024) which uses an innovative personal attack called Word Adversarial Forcing to Fail the Language model (WAFFLE) to show how input-adaptive multi-exit language models can be vulnerable to adversarial slowdown. The study demonstrates that WAFFLE drastically reduces the computational savings for early exit mechanisms, especially affecting more sophisticated methods that offer radical reductions in computation. What is more, they also examined subject-predicate disagreement and named entities changes as key factors for successful adversarial text crafting. The paper further evaluates alternative countermeasures, with the findings being that adversarial training doesn't work while input sanitation by conversational models like Conversational Generative Pre-trained Transformer (ChatGPT) works effectively. In quantitative terms, experiments show significant drops in model accuracies and increased computation overheads caused by WAFFLE-generated texts that are adversarial oriented. For example, when considering Multi-Genre Natural Language Inference(MNLI) data set, accuracy increases from 54% to 66% with a decrease from 11.5 average exit number to 7.5 exit number, whereas in Quora Question Pairs (QQP), accuracy rises from 56% to 80% with an exit number reduction from 9.6 to 7.1.

Also, as discussed in background section, TEXTFOOLER (Jin, Jin, Zhou and Szolovits 2020) is another framework that manages to generate adversarial examples against BERT while preserving the previously mentioned constraints. It first figures out important words in the text and substitutes those words with their synonyms until the prediction is altered. This is done while making sure that the context of the word and the semantics of the sentence is preserved. The attack model employed in the study reduces BERT's accuracy by approximately 5–7 times on classification tasks, marking an unprecedented level of vulnerability. This process highlights the importance of understanding and targeting specific words that have a substantial impact on the model's decision-making.

While TEXTFOOLER uses a black-box model where only the input and the output are available to the adversaries, there are situations where this isn't the case. Instances exist where adversaries successfully compromise a BERT-based API service across diverse benchmark datasets. Moreover, the susceptibility of BERT-based API services persists, even in situations where there is an architectural misalignment between the targeted model and the employed attack model, as highlighted in (He, Lyu, Xu and Sun 2021). In such scenarios, a more tailored way of knowing the importance of words is possible; we explore them in the upcoming section.

## 4 Security Issues and Exploitation

### 4.1 Model Vulnerabilities of BERT

#### 4.1.1 The Role of Word Importance

In BERT, the importance of words within a text is a key factor in determining the model's output or prediction. Some words carry more weight than others in influencing the final sentiment classification. Identifying these salient words is a complex task. BERT, being a deep neural network, processes text sequentially, considering the contextual relationships between words. Therefore, the importance of a word depends on its contextual significance. However, if we recognize these words, manipulating them can have a great impact on the model's behavior.

For example, in the sentence "The movie was incredibly joyful," the word "joyful" is pivotal in expressing a positive sentiment, while the words "movi3" and "was" contribute less to the overall sentiment. Therefore, understanding the importance of words in sentiment analysis is essential for crafting adversarial examples and manipulating the model's predictions.

#### 4.1.2 Gradients as a Tool:

Neural networks work by updating weights according to the gradients of loss function. In the context of BERT, gradients can represent vectors that capture how the model's loss function changes concerning variations in its input i.e., they can capture the influence of individual tokens on the model's prediction. When a text is processed by BERT, each word or token is assigned a numerical value based on its embedding in a high-dimensional vector space. Gradients help quantify how sensitive the model's predictions are to variations in these embeddings. Specifically, they measure how small changes in the embedding of a word impact the model's loss function.

Gradients offer precise insights into which words carry substantial influence in the sentiment classification. They can reveal which words, when altered, are likely to trigger changes in the model's output. This granularity and precision make gradients an invaluable tool for identifying word importance within BERT.

### 4.2 Exploitation

#### 4.2.1 Problem Statement

Before exploring the vulnerability of the model, we formally formulate our problem statement.

Given is a corpus of N tweets $X = \{x_1, x_2, . . ., x_N\}$ with their corresponding sentiments $Y = \{y_1, y_2, . . ., y_N\}$ where $y_i$ is the sentiment of $x_i$ and $y_i \in$ {positive, negative, neutral}, and a fine-tuned pre-trained BERT model $F: X \rightarrow (Y, C_{neg})$, which maps the input tweet X to the sentiment Y with confidence $C_{neg}$ representing the confidence of the tweet being negative.

For a tweet $x \in X$ such that $y = F(x) = (\text{negative}, c_k)$ and $c_k >= \delta$, i.e., a tweet with a negative sentiment and the confidence of it being negative greater than a threshold, we need to find a valid adversarial example $x_{adv}$ that conforms to the following requirements:

$$F(x_{adv}) = (y, c_{k'}) : (y' \neq y \text{ or } c_{k'} < \delta) \text{ and } Sim(x, x_{adv}) \geq \varepsilon$$

i.e., the adversarial example should either give non-negative sentiment or the confidence of it being negative should be below the defined threshold. Also, it should be semantically and syntactically similar to the given tweet.

where $Sim : x * x \rightarrow (0, 1)$ is a semantic and syntactic similarity function, $\varepsilon$ is the minimum similarity between the original and adversarial examples, and $\delta$ is the confidence threshold below which sentiment is ambiguous.

#### 4.2.2 Threat Model

We have a white-box scenario where the attacker has access to the fine-tuned pre-trained BERT model; they can access the model parameters and hence can exploit it. However, even without the fine-tuned model, a pretrained base BERT can still impose security threats as the model's learned representations can be similar even for a different dataset. This is suggested in the paper (He, Lyu, Xu and Sun 2021) which explores the vulnerabilities of BERT-based API services on model extraction attacks and adversarial example transfer attacks. The attacks highlight potential risks associated with the deployment of BERT-based models as services.

There are black-box frameworks to estimate importance of words, however we still choose a white-box scenario, specially we use gradients for measuring word importance for following reasons:

1) Fine-Grained Sensitivity: Gradients exhibit heightened sensitivity to even minor adjustments in input embeddings, making them exceptionally reliable for detecting word importance.
2) Precise Word Identification: Gradients offer precise insights into the impact of words on the loss function. This level of precision empowers our model to identify words that exert a substantial influence on model predictions.
3) Reduced Query Overhead with Gradients: In black-box models, the number of queries to the model to access importance of word scales linearly with the number of words in the tweet. Whereas gradient-based methods necessitate only two queries per tweet—once for the forward pass and once for the backward pass.

The steps are as follows:

**Step 0: Pre-Process**

The preprocessing pipeline incorporates essential steps for refining the input data. It leverages libraries such as pandas for data handling, spellchecker for correcting spelling errors, and contractions for expanding contractions in the text. The process involves handling slang words, correcting spelling errors, expanding contractions, and removing URLs to enhance the dataset's quality. The resulting preprocessed data is then saved for subsequent sentiment analysis tasks.

**Step 1: Find Important Words**

In this step, we analyze a given tweet to rank the importance of individual words. We leverage the BERT model's gradients; Gradients provide us with insights into how small changes in word embeddings can affect the model's loss. If you have a feature x, and $\partial L/\partial x$ (the gradient of the model's loss function with respect to x) is larger, it means that a small change in x ($\Delta x$) will result in a larger change in the model's loss ($\Delta L$). This is demonstrated for a quadratic model with a single parameter in Fig. 2 and Fig. 3.

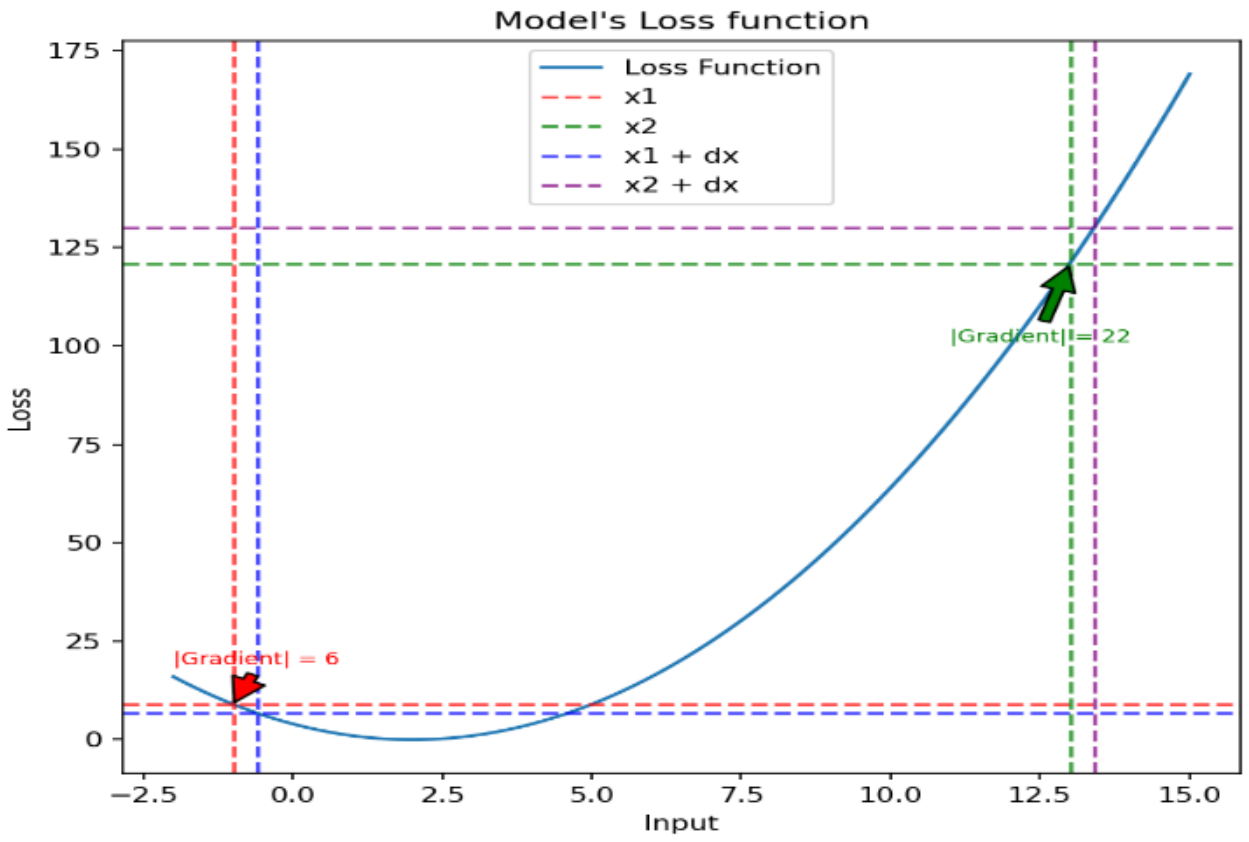

**Fig. 2** Model Loss against Inputs and their deviations, and respective gradients.

The importance ranking is essential for identifying words to target for replacement in the adversarial text generation process.

We use a Gradient tape to record the operations performed on tensors during a forward pass of the network and then use this record information to compute gradients during the backward pass. By doing this, since BERT works on tokens, we get the corresponding gradients for those tokens. To derive the importance of a word, we average out the gradients of the tokens that form the word. Finally, we normalize

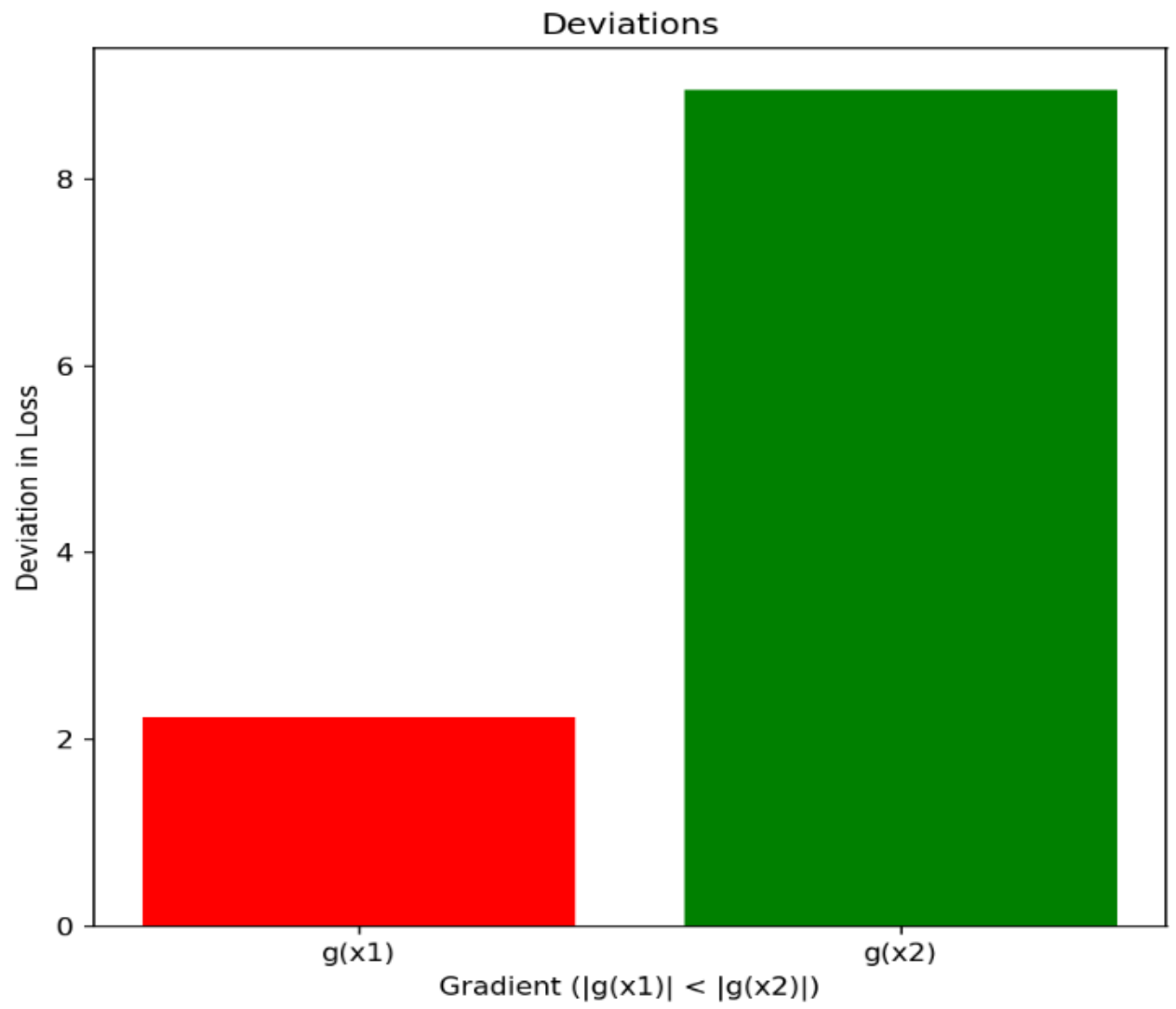

**Fig. 3** Deviation in Loss for small change in Input for different gradients

the importance scores, rank the words accordingly, and select the most important ones.

Mathematically,

Given is a tweet x of n words, $x = \{w_1, w_2, w_3, \ldots, w_n\}$, where each word is composed of one or more tokens $w_i = \{t_1, t_2, \ldots, t_m\}$, and L the loss function.

Let G(L, t) be the gradient of the loss function with respect to the token t, I'(w) be the importance of the word w and I(w) be the normalized importance of the word.

Tape: x -> y

$I'(w) = 1/n * \sum G(L, t_j)$ for $tj \in w$,

where $G(L, t) = \nabla L / \nabla t$

L is the loss function

$$I(w) = (I'(w) - \mu) / \sigma \quad (1)$$

where $\mu$ is the mean and $\sigma$ is the standard deviation of importance scores across all words.

We take words with $I(w) >= \theta$ where $\theta$ is a threshold for minimum importance of word. Let the important words be IW. Remove filter words from IW.

**Step 2: Select Feasible Candidates**

After ranking the words, we proceed to select feasible candidate words for replacement. The candidate selection process involves finding suitable replacement words that meet specific criteria:

a. Similar Semantic Meaning: Replacement words should have similar semantic meanings to the original words. This ensures that the overall message of the tweet remains consistent. We obtain this by using synonyms.
b. Same Part-of-the-Speech (POS): We only keep the synonyms with the same POS as the word to maintain the grammatical correctness.

For meeting both the criteria, we use Datamuse API (DataMuse. (n.d.). *Datamuse API)* and python's nltk library. For each word in IW, we now have candidates. Let $C(w_i)$ be the candidates for $w_i$ where $w_i \in IW$.

**Step 3: Generate Adversarial Texts**

For each $w_i \in IW$, we take candidates $C(w_i)$ as CANDIDATES and substitute the word in the original text x by a candidate to generate potential adversarial text $x_{adv.}$ We compare this with the original text using Universal Sentence Encoder (USE) (TensorFlowHub. 2023) and discard it if the context is changed. Otherwise, we test the adversarial text against our model. If the prediction is non-negative, we directly finalize this adversarial text. Else, we move on to the next candidate. If we deplete all candidates for a word, we take the candidate that has the least confidence for negative sentiment, replace the word in the text with this candidate and then move on to the next important word with the new text as the original text.

**Algorithm** Adversarial Attack on BERT

**Input:** Sentence example $x = \{w_1, w_2, ..., w_n\}$ with a negative sentiment, model F sentence similarity function Sim(·), sentence similarity threshold ε, and synonyms from Datamuse API.

**Output:** Adversarial example Xadv

Initialization: xadv ← x

**for** each word wi in x **do**

Compute the importance score I(wi) via Eq. (1)

**end for**

Create a set w of all words wi ∈ x sorted by the descending order of their importance score I(wi).

Remove words from w if I(wi) < θ.

Filter out the stop words and punctuations in w.

y, neg_conf ← F(xadv)

**for** each word wj in w **do**

Initiate the set of candidates C(wj) by extracting the top N synonyms Datamuse API.

// Only take candidates with same part-of-speech as the word

C(wj) ← POSFilter(C(wj))

**for** candidate in C(wj) **do**

x' ← Replace wj with candidate in xadv

// Replacements are only valid if the resulting sentence is similar enough to the original sentence

**if** Sim(x', xadv) > ε **then**

Add candidate to the set FINCANDIDATES

// Predict the label and negative sentiment confidence using the model

y', neg_conf' ← F(x')

// If the label is non-negative, generation successful.

**if** y' != y **then**

**return** x'

**end if**

**end for**

min_neg_conf ← min negative confidence(FINCANDIDATES)

c ← candidate with min_neg_conf

// Replace word with candidate having minimum negative confidence

xadv ← Replace wj with c in xadv

**end for**

// If negative sentiment confidence is smaller than threshold, generation successful

**if** min_neg_conf < δ **then**

**return** xadv

**end if**

**return** None

**4.2.3 Analysis Model and Dataset Description**

In our analysis, we use a pre-trained BERT model obtained from Hugging Face. This Sentiment Analysis BERT Model (Hugging Face. 2023) has been trained for sentiment classification. To fine tune it for our tweet classification task, we train in on a tweet dataset from Kaggle (M. Yasser H, "Twitter Tweets Sentiment Dataset,"), comprising 27,481 rows of Twitter tweets all mapping to either of the three labels, positive, neutral, or negative. The dataset includes columns such as textID, serving as a unique identifier for each piece of text, the tweet text itself, and the corresponding sentiment label indicating the general sentiment of the tweet. The code is openly accessible on GitHub (AKIL003 BERT-Attack 2023).

## 5 Analysis and Discussion

### 5.1 Result

In our experiment, we successfully generate a total of 5046 adversarial examples, representing 64.86% of the total tweets initially classified with negative sentiment by the fine-tuned pre-trained BERT model. This drops the accuracy of the fine-tuned model from 80.33% to 70.29%, which might not seem as impressive when compared to the literature, but it is because our objective isn't to reduce the accuracy, rather it is to remain as stealthy as possible while performing targeted misclassification. On average, 2 words are replaced per successful attempt, highlighting the effectiveness of the attack in altering the model's predictions. The minimum semantic and syntactic similarity threshold (ε) was set at 0.9, ensuring that generated adversarial examples retained meaningful correspondence with the original tweets. Additionally, to identify instances where sentiment became ambiguous, a confidence threshold (δ) was utilized, with a value of 0.5. These results underscore the vulnerability of sentiment analysis models, emphasizing the importance of robust defenses against adversarial attacks on natural language processing tasks.

**Table 1** Adversarial Text Examples

| Original Text | Adversarial Text |
|---|---|
| finally sunny days and i`m too sick to go outside and play. | finally sunny days and i`m too infirm to go outside and play. |
| i`m trying to find the driver for my microsoft ex-3000 webcam and can`t find it anywhere! anyone have any links? | i`m hoping to find the driver for my microsoft ex-3000 webcam and can`t find it anywhere! anyone have any links? |
| stupid bipolar weather ruined my day off | rediculous manic weather obliterated my day off |
| _re234 haha yup. but still have a terrible headache and super swollen and puffy eyes! i do not think i am going out today.ugh! | _( haha hehe. but still have a atrocious migraine and super swollen and puffy eyes! i do not think i am going out today.ugh! |
| that is pretty bad quality and probably the worst pic you posted till date | that is awesomely iffy quality and probably the worst pic you posted till date |

In Table 1, we present few examples of adversarial texts generated during our experiment, illustrating the potency of the targeted gradient attack in altering sentiment predictions. These examples underscore the subtlety and effectiveness of word replacements in influencing the model's output, providing tangible evidence of BERT's vulnerability to manipulation.

### 5.2 Discussion

Our investigation underscores the influence of specific words on sentiment predictions, with gradients providing valuable insights into individual word impact on the overall model output. BERT's contextual understanding, while a strength, becomes a vulnerability when it comes to the targeted manipulation of influential words. This susceptibility raises concerns about the robustness of BERT in the face of nuanced adversarial attacks.

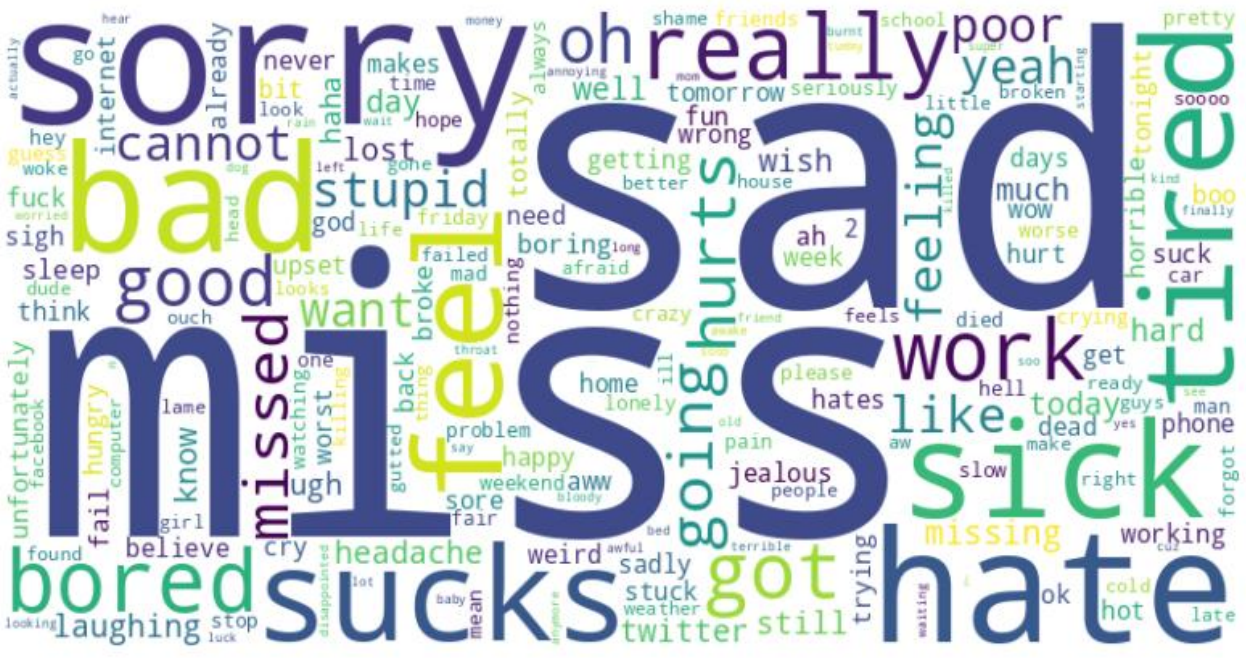

**Fig. 4** Word cloud of the most frequent important words.

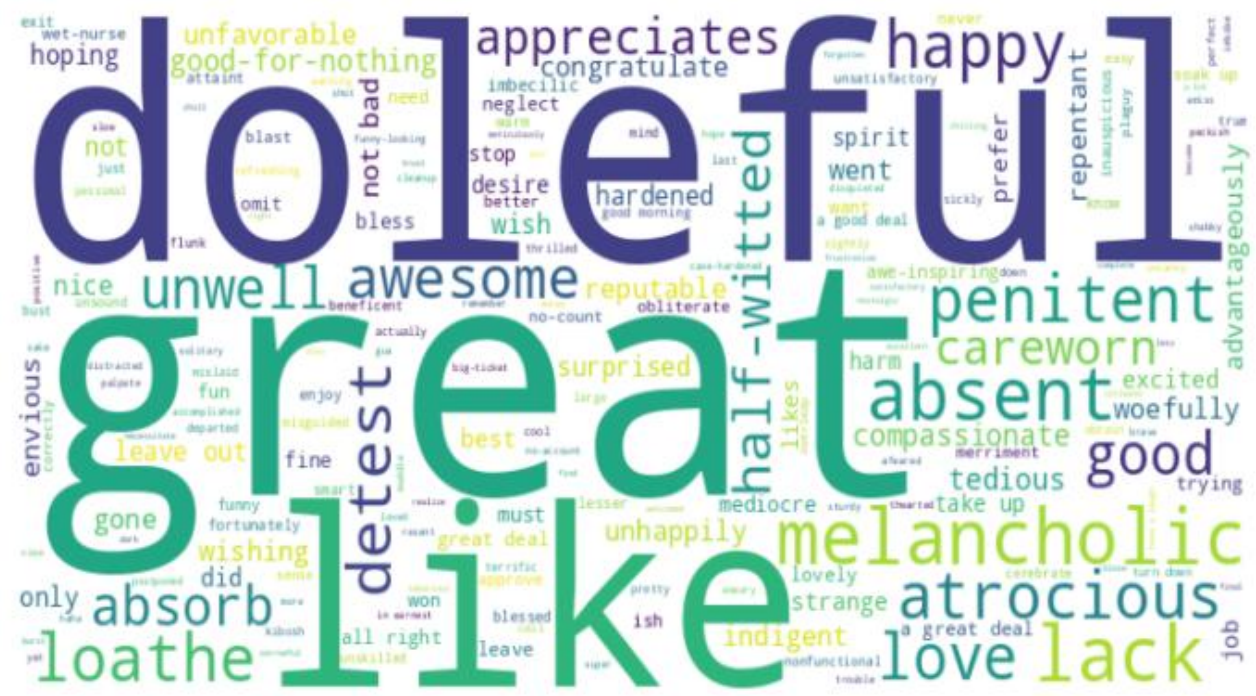

**Fig. 5** Word cloud of the most frequent words that triggered the change in prediction.

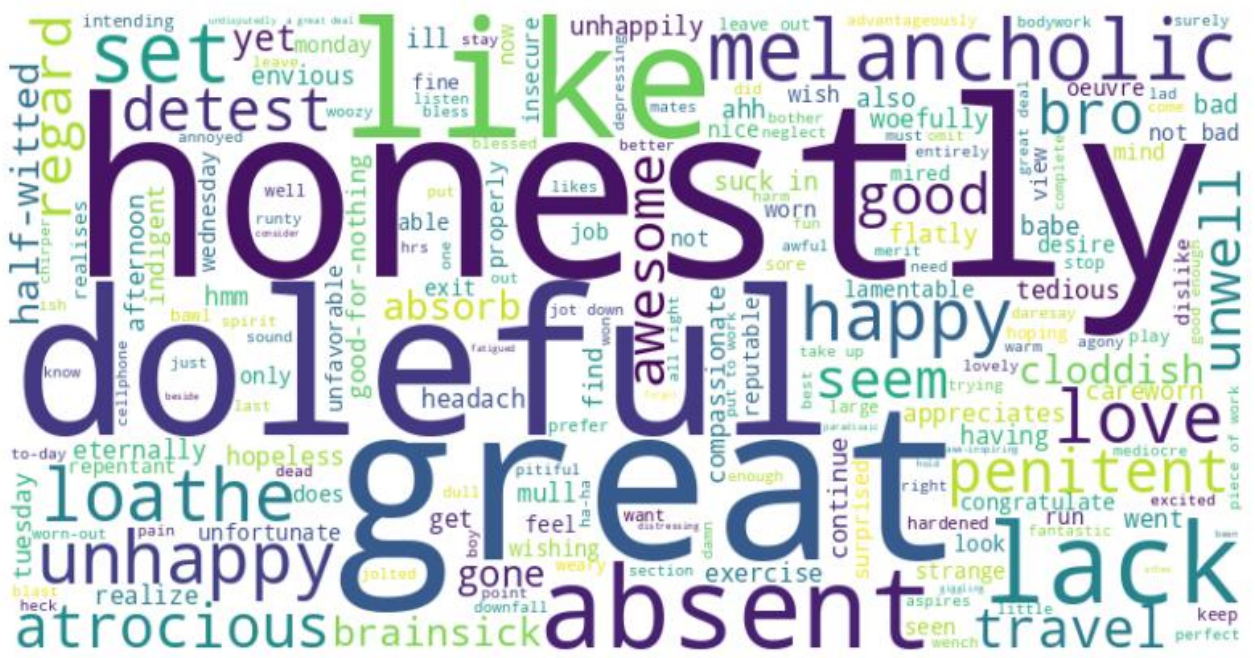

**Fig. 6** Word cloud of the most frequent words that replaced the keywords.

Our presentation of three key word clouds (Fig. 4, 5 and 6) serves as visual evidence supporting our claims regarding the dynamics of sentiment analysis with BERT. Fig. 4, depicting the most frequent words pivotal for prediction, aligns with our assertion that certain words carry more significance in influencing sentiment outcomes. This visual representation underscores the model's focus on specific terms. In parallel, Fig. 5, and Fig. 6, showcasing words that substituted the identified words and triggered specific predictions, depicts the direct relationship between language nuances and sentiment analysis outcomes, reinforcing our claim that gradients can point to important words and replacing these identified words can indeed alter predictions. Together, these visual representations offer compelling support for the central tenets of our investigation.

Examining the broader implications of our findings, the susceptibility of BERT to targeted gradient attacks raises significant concerns which we would discuss in security implications.

## 6 Security Implications

Our examination of the highly regarded fine-tuned BERT model reveals substantial concerns with tangible consequences. Despite BERT's reputation for robustness, the success of a targeted gradient attack and the generation of adversarial examples underscore the nuanced risks associated with misclassifying sentiments in natural language processing systems, emphasizing the need for careful consideration.

**6.1 Misinformation and Trust Erosion:** While BERT is widely recognized for its robustness, the ability to manipulate sentiment classifications raises concerns about the potential spread of false information, highlighting the intricacies of sentiment analysis. Adversarial examples that tweak the perceived sentiment of tweets could propagate misleading narratives, impacting public opinion and posing challenges for users who place their trust in BERT's sentiment analysis for decision-making.

**6.2 Security Challenges on Social Platforms and Campaign Vulnerability:** Social media platforms utilizing BERT for sentiment analysis in content moderation and user experience improvement face specific security challenges. The success of adversarial examples may pose a risk to content filters, potentially impacting moderation and allowing the spread of inappropriate or harmful content. Moreover, political or marketing campaigns relying on BERT's sentiment analysis could be vulnerable to manipulation, raising questions about the robustness of sentiment analysis applications in real-world scenarios.

### 6.3 Ethical Concerns Amplified

The ethical implications of manipulating sentiment analysis models reveal worries about user privacy and fairness. Despite the widespread trust in BERT, adversarial attacks, especially if they disproportionately impact certain user groups, intensify ethical considerations about the fairness and unbiased nature of BERT's sentiment analysis applications.

## 7 Conclusions

Our investigation employs a targeted gradient attack methodology to explore the vulnerability of sentiment analysis models, focusing specifically on the fine-tuned BERT model. Through this approach, we can successfully generate adversarial examples, manipulating sentiment classifications while maintaining semantic and syntactic similarity. Notably, an average of ~2 words are replaced per successful attempt, highlighting the effectiveness of the attack.

Our primary objective was fulfilled as we demonstrated the susceptibility of the BERT model to misclassification, emphasizing the need for robust defenses in natural language processing systems. Beyond technical intricacies, our findings underscore the real-world implications of potential misinformation propagation, challenges to user trust, and security issues on social platforms. As we uncover the nuanced risks tied to BERT's sentiment analysis, our study contributes to the ongoing discourse on securing automated systems and maintaining the integrity of sentiment analysis applications in the digital landscape.

**Statements and Declarations**

**Conflict of interest**

The authors declare no conflict of interest. The funding agency had no role in the design of the study; in the collection, analyses, or interpretation of data; in the writing of the manuscript, or in the decision to publish the results.

**Funding Statement**

This research received no specific grant from any funding agency in the public, commercial, or not-for-profit sectors. The authors conducted this study independently and without external financial support.